\documentclass[10pt,twocolumn,letterpaper]{article}

\usepackage{times}
\usepackage{amsmath}
\usepackage{amssymb}

\usepackage[table]{xcolor}
\usepackage{pgfplotstable}

\usepackage[pagebackref=true,breaklinks=true,letterpaper=true,colorlinks,allcolors=blue!60!black,bookmarks=false]{hyperref}

\begin{document}

\title{Plane Pair Matching for Efficient 3D View Registration}

\author{Adrien Kaiser \textsuperscript{1,2}\\
\and
Jose Alonso Ybanez Zepeda \textsuperscript{2}\\
\and
Tamy Boubekeur \textsuperscript{1}\\
}

\date{\normalsize \textsuperscript{1} Telecom Paris, Institut Polytechnique de Paris, France\\
\normalsize \textsuperscript{2} Fogale Nanotech, Le Kremlin Bicetre, France\\
}

\maketitle

\begin{abstract}
   
We present a novel method to estimate the motion matrix between overlapping pairs of 3D views in the context of indoor scenes.
We use the Manhattan world assumption to introduce lightweight geometric constraints under the form of planes into the problem,
which reduces complexity by taking into account the structure of the scene.
In particular, we define a stochastic framework to categorize planes as vertical or horizontal and parallel or non-parallel.
We leverage this classification to match pairs of planes in overlapping views with point-of-view agnostic structural metrics.
We propose to split the motion computation using the classification and estimate separately the rotation and translation of the sensor, using a quadric minimizer.
We validate our approach on a toy example and present quantitative experiments on a public RGB-D dataset, comparing against recent state-of-the-art methods.
Our evaluation shows that planar constraints only add low computational overhead while improving results in precision when applied after a prior coarse estimate.
We conclude by giving hints towards extensions and improvements of current results.

\end{abstract}

\section{Introduction}
\label{sec:introduction}

This paper focuses on an essential task in the process of recording the real world in 3D, namely the registration of multiple 3D views.
We aim to introduce geometric constraints under the form of planes into the registration problem, in order to reduce complexity and take into account the structure of the scene.
We present methods to match planes between overlapping 3D views in \autoref{sec:matching} and use these matches to estimate the motion matrix between the views in \autoref{sec:motion}.
Experiments of 3D view registration based on detected planes are shown in \autoref{sec:evaluation}.

The specificity of our approach is the structure analysis we implicitly perform
when categorizing planes given their orientation in the scene and comparing them by pair instead of one by one.
In addition, our plane-based motion estimation leverages this classification to separately estimate the different degrees of freedom of the sensor
based on plane orientation matching and \emph{quadric error} minimization.
Existing methods that ignore the scene structure are more prey to confusion between geometrical elements
and require a certain amount of constraints and their configuration in order to estimate the full camera motion.

\subsection{Context and Motivation}
\label{sec:intro_context}

The low amount of structural and high level information contained in a single depth view can be increased by aggregating several views.
However, this requires knowledge of the relative position and orientation of the sensor while capturing these views.
The problem of image registration, while studied for several decades, gets more complex when considering 3D data, as shown in \autoref{sec:relatedwork}.
In this study, we show that estimating the motion between different 3D views is mostly carried out using local 3D descriptors such as \emph{fast point feature histograms (FPFH)} \cite{rusu2009fast},
or 2D descriptors using the 3D information of the depth component, when color image is available.
Instead, our approach makes use of detected planes in overlapping 3D views to infer relative position and orientation of sensors.

In order to stay as general as possible and as we cannot assume having control over the acquisition of the views,
we consider registration of 3D views by single pairs that overlap between roughly 20\% and 80\%
in \emph{intersection over union} of the 3D geometry e.g., image surface or 3D point locations.
Empirically, these values give sufficient overlapping features while keeping challenging geometrical differences.
We do not aim to use any kind of global optimization as we only consider two views at a time and not the full acquisition.
Our algorithm, being based on detected planes, is particularly suited for indoor environments composed of multiple horizontal and vertical planar structures,
such as floors, ceilings, walls, doors or tables.
More specifically, scenes that follow the \emph{Manhattan world} orientation assumption \cite{coughlan1999manhattan} allow better performance.

The inference of sensor motion from the scene structure gives robustness to variability in indoor scenes where small objects might often be moved.
Hence, such a registration of scenes acquired at different times is robust to movements of objects, which could confuse local descriptors.

\subsection{Overview}
\label{sec:intro_overview}

\begin{figure*}
\centering
    \includegraphics[width=\textwidth]{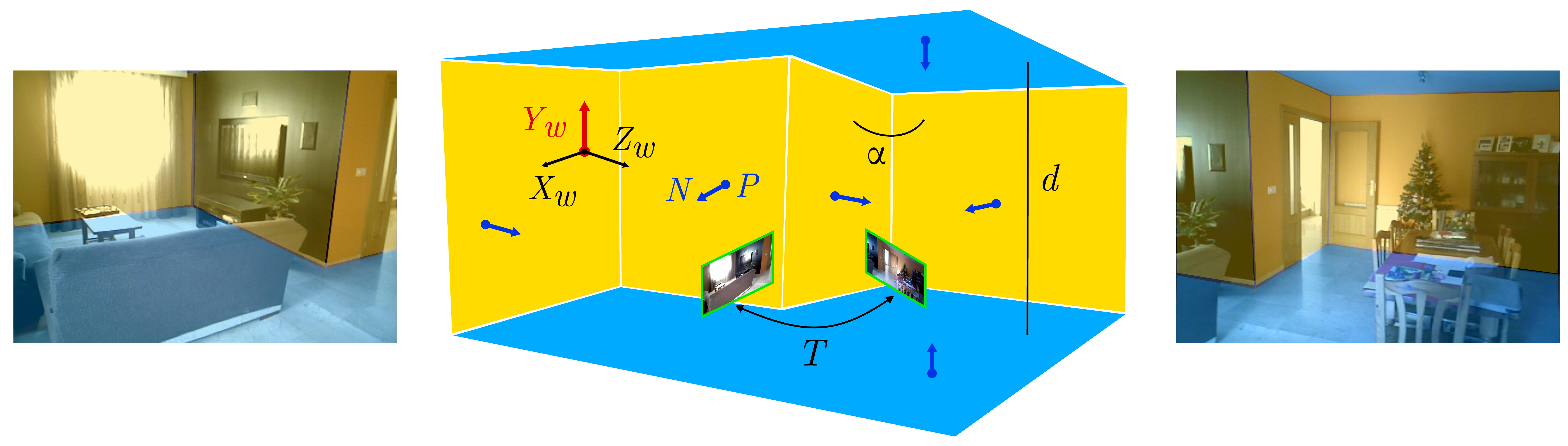}
\caption
{Overview of the plane-based registration of two RGB-D views.
Two RGB-D views are represented by their color component with planar areas overlaid (left and right).
In this example, the observed indoor scene (middle) is composed of vertical planes, the walls (orange) and horizontal planes, the floor and ceiling (light blue).
A plane is represented by its normal vector $\vec N$ and a 3D point $P$ at its surface (dark blue arrows and circles).
The scene was captured by the views at the locations shown in green frames when the sensor moved of a motion $T \in \mathbb{R}^{4 \times 4}$.
Using our analysis of the scene, we define the world local frame $\vec X_w, \vec Y_w, \vec Z_w$ where $Y_w$ is the \emph{Up} vector aligned with the \emph{gravity}.
Our method is able to recover the sensor motion using plane arrangements constraints such as relative angles $\alpha$ and distances $d$.}
\label{fig:intro_overview}
\end{figure*}

Our goal is to leverage the scene structure to estimate the motion $T$ of a sensor capturing an indoor scene.
We decompose the scene into different plane arrangements, e.g. horizontal and vertical, using our probabilistic framework
based on geometric characteristics of planes, such as relative angles $\alpha$ and distances $d$.
By defining a local coordinate frame, we are able to reduce the dimensionality of the computation and separately estimate the degrees of freedom of the sensor.
\autoref{fig:intro_overview} is a representation of the elements involved in our plane-based registration procedure.

\subsection{Contributions}
\label{sec:intro_contributions}

We present different strategies to match planes between 3D views and estimate the relative transformation based on the following contributions:

\begin{itemize}
    \item the definition of a probabilistic framework to analyze the scene structure and separate different arrangements of planes, e.g. horizontal and vertical;
    \item the comparison of planes by pairs instead of single matches to reduce the complexity of the search;
    \item the definition of multiple heuristics based on geometry and appearance to prevent wrong matches;
    \item a quadric minimizer between multiple vertical planes to estimate the horizontal translation of the sensor;
    \item the estimation of a prior motion matrix to gracefully degrade to the state-of-the-art in cases where not enough plane information is available.
\end{itemize}

\section{Related Work}
\label{sec:relatedwork}

\subsection{Pairwise RGB-D Frames Registration}
\label{sec:related_framepair}

This section focuses on pairwise registration of overlapping views of a scene acquired by an RGB-D sensor.
The input is composed of two single RGB-D images of an indoor scene sampling overlapping objects.
The goal is to estimate the rigid transformation matrix that transforms the first scan into the second scan.
It also corresponds to the motion of the sensor between the two acquisitions.
In a recent survey, Morell-Gimenez et al. \cite{morell2018survey} divide these methods into two categories.
First, the construction of sparse feature descriptors at 2D or 3D point locations and their matching using RANSAC \cite{fischler1981random}.
Second, the dense registration methods pioneered by \emph{Iterative Closest Point (ICP)} \cite{besl1992method}
and applied to depth data through the \emph{KinectFusion} \cite{newcombe2011kinectfusion} line of work.

Here, we only detail methods dedicated to RGB-D frames, but one may also use point cloud methods,
as described in \autoref{sec:related_pointcloud}, to register depth maps.
Most methods aim at matching 3D locations between the two views and then compute the motion matrix by minimizing their Euclidean distance in the \emph{least squares} sense \cite{umeyama1991least}.
In order to acquire corresponding 3D locations, a first range of methods uses the color component and known 2D keypoint detectors and descriptors.
After matching 2D descriptors in the color component such as SIFT \cite{lowe1999sift}, SURF \cite{bay2006surf} or ORB \cite{rublee2011orb},
the corresponding 3D locations can be directly read from the depth component.
While the use of known 2D descriptors opens the range of available tools, the color component is sensitive to illumination changes,
which can confuse the matching of such features if acquired at different times.

Recent methods leverage the global availability of RGB-D reconstruction datasets to learn local 3D feature descriptors.
In particular, \emph{3DMatch} \cite{zeng20163dmatch} descriptors are learned through a self-supervised volumetric convolutional neural network.
Ground truth correspondences are acquired using existing reconstructed models of RGB-D datasets
and the definition of volumetric patches allows computing strong local descriptors that can be matched using RANSAC.
While the \emph{3DMatch} descriptors show robustness in a variety of scenarios, it is important to note the high requirements of the method in terms of hardware and execution time.

\subsection{Pairwise Point Cloud Registration}
\label{sec:related_pointcloud}

In this section, we present more general registration methods applied to 3D point sets.
On one hand, efficient 3D descriptors can be matched between point sets and the transformation matrix can be estimated as for RGB-D frames.
On the other hand, the family of \emph{iterative closest point (ICP)} variants allows accurate registration at the cost of multiple iterations over the full dataset.

\subsubsection{3D Descriptors}

Visual descriptors aim at describing features in visual data in terms of shape, color or texture.
By analyzing the data at specific locations and their neighborhoods, they build a list of local characteristics to uniquely identify parts or elements.
Here, we present the most used 3D descriptors used in the process of pairwise point cloud registration.
For a comprehensive study, we point the reader to the 2016 survey from Guo et al. \cite{guo2016comprehensive}.

In 1999, Johnson and Hebert present the \emph{spin images} \cite{johnson1999using} with the goal of recognizing 3D objects.
At selected 3D oriented point locations, a local image plane is defined using cylindrical coordinates.
Projecting nearby points onto the virtual image plane allows aggregating local geometry information into a 2D, simple to compare image grid.

In 2008, Rusu et al. \cite{rusu2008aligning} define the \emph{point feature histogram (PFH)} descriptor to locally encode the geometrical properties of a 3D point's neighborhood.
By computing local deviations in surface orientation, they build a high dimensional histogram that describes well the local curvature and its variations.
Although, as the base information used is the surface normals, the robustness of the method highly depends on their quality.

In 2008, Aiger et al. present a method to match congruent groups of 4 points in \emph{4PCS} \cite{aiger20084points}.
By designing local metrics for a group of 4 points and analyzing their geometry, they are able to reach a high level of description even in the presence of noise and outliers.
This descriptor was then optimized by Mellado et al. in \emph{Super4PCS} \cite{mellado2014super4pcs} to reach linear complexity.

In 2010, Drost et al. present a new local geometric descriptor called \emph{point pair features (PPF)} \cite{drost2010model}.
A \emph{PPF} is defined for two oriented points to describe their relative position and orientation and an accumulator space allows fast comparison between point pairs.
A recent paper presents \emph{PPFNet} \cite{deng2018ppfnet}, a learning approach to the detection and estimation of \emph{PPF} descriptors.

\subsubsection{Iterative Closest Point}

The \emph{iterative closest point (ICP)} methodology was first described by Besl and McKay in 1992 \cite{besl1992method}
with the goal of registering general 2D and 3D geometric objects of multiple representations.
\emph{ICP} is widely used to register 3D point clouds and has seen numerous extensions.
The first one was its combination with the point-to-plane distance defined by Chen and Medioni \cite{chen1991object},
by replacing the original point-to-point metric by the distance between a 3D point and the tangent plane at its corresponding oriented point.

\emph{Generalized ICP} by Segal et al. \cite{segal2009generalized} as well as Viejo et al. \cite{viejo20083dmodel}
define a \emph{plane-to-plane ICP} by considering both source and target point clouds as oriented, thus having tangent planes at each location.
Finally, \emph{Sparse ICP} by Bouaziz et al. \cite{bouaziz2013sparseicp} is an efficient variant of the original method, where outliers of the transformation are detected and removed during the process.

For more details on \emph{ICP} variants, a study was published by Rusinkiewicz and Levoy \cite{rusinkiewicz2001efficient}.
More recently, Bellekens et al. \cite{bellekens2014survey} also compared \emph{ICP} variants to registration based on \emph{principal component analysis} and \emph{singular value decomposition}.

\subsection{Offline Global Registration}
\label{sec:related_global}

While we have seen state-of-the-art methods to rigidly register pairs of overlapping RGB-D frames or point clouds,
we now focus on the use of multiple images to register them together into a global model of the scene.
Processing the full acquired collection of frames allows building a more complete model where missing data from some views is filled by other views.
We consider offline global registration, in opposition to online reconstruction that provides live interactive generation of a scene representation \cite{endres20143dmapping, newcombe2011kinectfusion}.

Most methods first consider an initial pairwise transformation as previously described.
Then, different strategies allow aggregating the information from all registered frame pairs into the global model.
The \emph{structure-from-motion} method of \emph{SUN3D} \cite{xiao2013sun3d} first registers a list of RGB-D frames with each other
using 2D keypoint descriptors associated with a RANSAC framework.
Then, they define an energy based on geometry, appearance and semantics
which is iteratively solved in order to get all frames into a globally consistent coordinate frame.

Methods by Choi et al. \cite{choi2015robust} and Zhou et al. \cite{zhou2016fast}, whose implementations are both available in the \emph{Open3D} library \cite{zhou2018open3d}, use the \emph{fast point feature histogram (FPFH)} descriptor to estimate initial pairwise alignments of partial point clouds.
Then, a RANSAC framework allows discarding wrong initial alignments and a global bundle adjustment pass leads to high quality reconstruction results.

\subsection{Plane-based Registration}
\label{sec:related_planebased}

When looking to localize acquired parts of indoor scenes with each other, several methods make the observation that they are mostly composed of planar elements.
Hence, introducing planar constraints into both pairwise and global registration improves robustness and reduces the drift due to data accumulation.
In addition, using planar geometric features can improve localization in low textured areas of the scene.

In 2010, Pathak et al. \cite{pathak2010fast} define a consistency-based framework to match extracted local plane features between multiple 3D views.
By introducing the plane parameters into a \emph{SLAM} update step, they obtain regular and accurate indoor scene reconstruction from a depth sensor mounted on a mobile robot.

In a similar fashion, Dou et al. \cite{dou2012exploring} detect large planes in indoor scenes and modify the RANSAC matching step to handle both points and planes.
The generation of multiple hypothetic plane matches and their disambiguation using plane pairs and relative angles improves the robustness of the method.
Again, the extension of \emph{bundle adjustment} to planes greatly improves the accuracy of the reconstruction results.

Forstner and Khoshelham \cite{forstner2017efficient} assume prior knowledge of large matching planes in overlapping 3D views
to define a plane-to-plane metric that is minimized to get the relative sensor positions and orientations.
By explicitly modeling the uncertainty of detected planes, they leverage planar constraints and provide multiple formulations and solutions to the registration problem.
Their method is fast and accurate to register pairs of frames, however gets confused and slower when multiple frames are simultaneously processed.
The prior requirement of matched planes, which itself is a complex task, is a limit to the automation of this technique.

Halber and Funkhouser \cite{halber2017finetocoarse} define a \emph{fine-to-coarse} registration framework that iteratively aggregates local to global planar features matched in subsets of the scene.
Enforcing planar regularity while preserving local features fixes drifting issues appearing in regular optimization frameworks.
In addition, the local to global aggregation reduces the sensitivity to errors appearing in the frame-to-frame registration based on \emph{SIFT} and \emph{RANSAC}.
However, as an offline method, it can not be applied to live registration because of the requirement for global structural information.

More recently, Shi et al. presented \emph{PlaneMatch} \cite{shi2018planematch}, a learning approach to planar feature matching and registration in RGB-D frames.
They predict local and global patch coplanarity in different RGB-D views of a scene
and aggregate all coplanarity constraints as well as point correspondences into a robust optimization framework that achieves state-of-the-art results.
The predictions are mostly accurate and of various nature, even with wide baselines, thanks to the large amount of training data.
Although, as they rely on color, depth and normal information to match planar patches,
the lack of discriminative features e.g. flat or low textured areas, can sometimes lead to false positive matches disturbing the global registration.
In addition, the use of a \emph{neural network} requires high performance hardware and leads to high computation times.

\section{Plane Matching}
\label{sec:matching}

In this section, we present strategies to match planes between two views of a scene, represented as a depth map and more generally as a point cloud.
The first step is to estimate the parameters of the planar elements of the scene in both views.
This can be done through different methods, detailed in a recent survey \cite{kaiser2018survey}.
In our experiments, we use the \emph{RANSAC-based} plane detection of Schnabel et al. \cite{schnabel2007efficient}.
At the end of this step, we have a list of detected planes, their associated parameters and inlier point positions.
In case we have a prior coarse estimation of the motion between the views, we can track the planes instead of detecting and matching them, as explained in \autoref{sec:matching_tracking}.

In the general case where we have no prior knowledge of the motion between the views, we first classify planes following absolute and relative geometry rules,
as explained in \autoref{sec:matching_horizvert}.
In particular, we separate horizontal and vertical planes relatively to the gravity orientation and classify them as parallel or non-parallel.
Grouping planes following these arrangements allows reducing the complexity of the matching problem.
Then, the matching is done by considering planes by pairs and comparing their relative orientation or distance in the two views, as explained in \autoref{sec:matching_association}.

\subsection{Plane Tracking}
\label{sec:matching_tracking}

A straightforward way to match planes between views is to use a prior transformation matrix between the views,
that could be estimated through any registration method presented in \autoref{sec:relatedwork}.
Then, we can run any plane detection method on the first view, apply the prior motion matrix to their parameters,
and check if samples of the planes are present in the second view.
If the number of inliers that are close enough to the planes in the new view is high enough,
then the plane is considered as seen in the new frame and its parameters can be refined using the new inlier set.

\subsection{Plane Arrangements}
\label{sec:matching_horizvert}

In the following, we will consider planes and pairs of planes as belonging to geometric categories with relation to the orientation of the scene, such as:

\begin{itemize}
	\item horizontal plane, i.e. orthogonal to a reference direction;
	\item vertical plane, i.e. parallel to a reference direction;
	\item pair of parallel planes;
	\item pair of non-parallel planes.
\end{itemize}

\subsubsection{Reference Direction}

For a given scene represented as an RGB-D frame or point cloud, we consider the reference direction to be the gravity vector in the sense of \emph{Newton's law} of universal gravitation in physics.
We motivate this choice after observing that most of the components of indoor scenes are either orthogonal or parallel to the gravity vector.
In particular, this direction has the advantage of being consistent in the scene and among the objects composing it, from any point of view and representation.

For the rest of this paper, we consider that the direction of the gravity is known and can be acquired by either one of these means:
\begin{itemize}
	\item identification of a near horizontal plane among the detected planes;
	\item inertial device such as an accelerometer, whose orientation in the coordinate frame of the data is known;
	\item pre-computed and available in a file for loading.
\end{itemize}

\subsubsection{Plane Classification}

\begin{figure*}
\centering
    \includegraphics[width=0.8\textwidth]{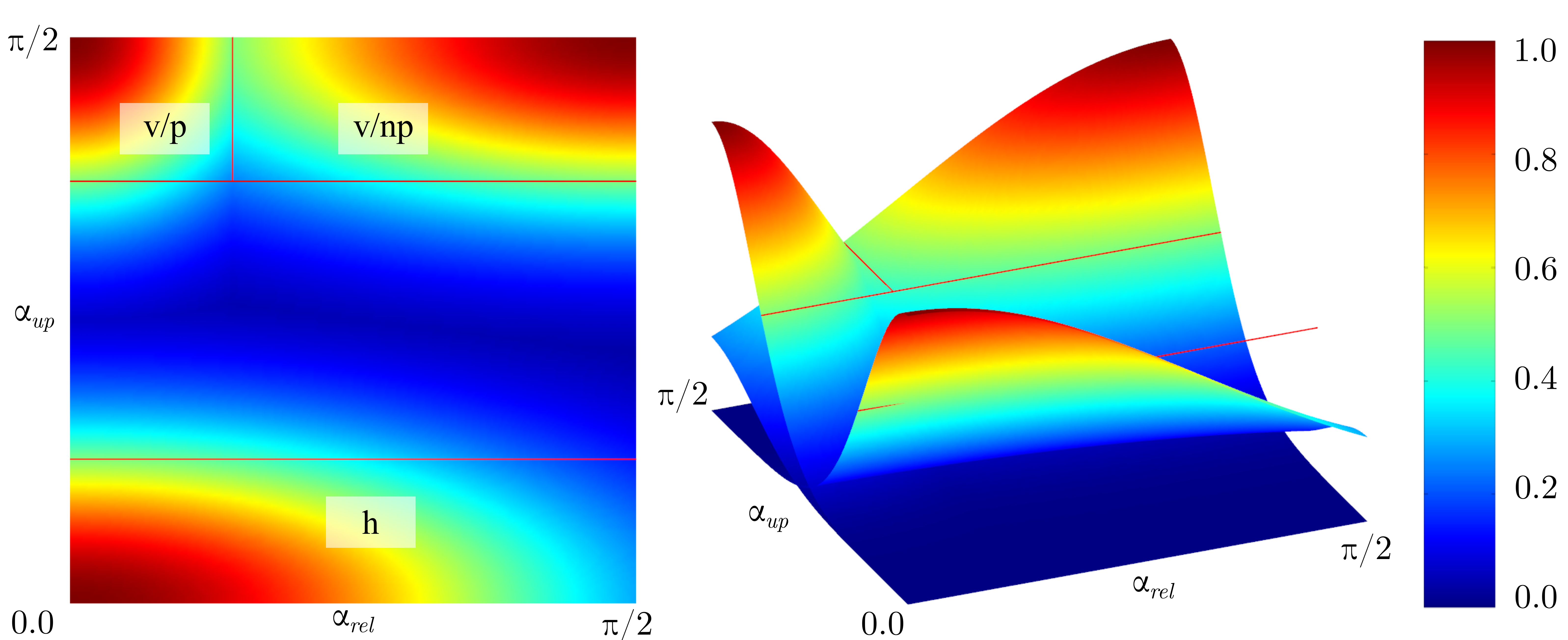}
\caption
{
Detection of plane arrangements using \emph{Gaussian} distributions.
The category of the plane or pair of planes is given by its highest probability.
Categories are horizontal (h), vertical parallel (v/p) and vertical non-parallel (v/np).
In this illustration, the y axis is the angle $\alpha_{up}$ of a plane with relation to the reference Up vector.
The x axis is the relative angle $\alpha_{rel}$ between two planes.
Control over the angle thresholds is given by the standard deviation values of the Gaussian functions.
The red lines show the actual angle thresholds in radians at a probability of 0.5.}
\label{fig:matching_horizvertgauss}
\end{figure*}

In order to classify planes as horizontal or vertical and parallel or non-parallel, we define the following probabilistic framework.
\autoref{fig:matching_horizvertgauss} shows visual representation of the plane classification probabilities.
For a plane of angle deviation $\alpha_{up} = \arccos(|\vec N . \vec Y_w|)$ with the reference direction $\vec Y_w$ and a pair of planes $(i,j)$ of relative angle deviation $\alpha_{rel} = \arccos(|\vec N_i . \vec N_j|)$, the class of the plane is defined by \autoref{eq:matching_horizvertclass}.

\begin{equation}
\begin{array}{l}
k(\alpha_{up}, \alpha_{rel}) \\ = \arg\max_k ~ g(\alpha_{up}, \mu^k_{up}, \sigma^k_{up}) ~ g(\alpha_{rel}, \mu^k_{rel}, \sigma^k_{rel})
\end{array}
\label{eq:matching_horizvertclass}
\end{equation}

Here, $g(\alpha, \mu, \sigma)$ is the value at $\alpha$ of a Gaussian function centered on $\mu$ with standard deviation $\sigma$, as shown in \autoref{eq:matching_gauss}.

\begin{equation}
g(\alpha, \mu, \sigma) = e^{\frac{- (\alpha - \mu)^2}{2\sigma^2}}
\label{eq:matching_gauss}
\end{equation}

The probability distributions $g(\alpha)$ for the three categories are centered at the following values:

\begin{equation}
\begin{array}{l l l}
\textrm{horizontal} & \mu_{up} = 0 & \mu_{rel} = 0 \\
\textrm{vertical parallel} & \mu_{up} = \pi/2 & \mu_{rel} = 0 \\
\textrm{vertical non-parallel} & \mu_{up} = \pi/2 & \mu_{rel} = \pi/2
\end{array}
\label{eq:matching_horizvertmeans}
\end{equation}

The value of $\sigma$ for each category is set to control the values of the angle thresholds by arbitrarily fixing the probability at 0.5.
In practice, we evaluate the standard deviation value $\sigma$ for each Gaussian function using an angle threshold value $\alpha_{thresh}$ and the fixed probability threshold 0.5, as shown in \autoref{eq:matching_horizvertgaussstd}.

\begin{equation}
\sigma(\alpha_{thresh}) = \sqrt{\frac{- (\alpha_{thresh} - \mu)^2}{2 ~ \log(0.5)}}
\label{eq:matching_horizvertgaussstd}
\end{equation}

In that formulation, the $\sigma$ values represent uncertainties associated with the plane classification, as they model the extent of the higher probability values over the angle domain.
Tuning these values gives control over the angle thresholds as well as the uncertainty of the classification.

\subsection{Plane Association}
\label{sec:matching_association}

We now consider that we have a list of planes categorized as horizontal or vertical and parallel or not.
In order to match planes that correspond to the same part of the actual scene, we will use this classification to reduce the amount of possible matches.
Hence, we first group horizontal and vertical planes together.
While we could simply try to match planes with each other in both views,
we choose to consider planes as pairs and not as single planes, in order to further reduce potential wrong matches.

\subsubsection{Plane Pairs Generation}
\label{sec:matching_association_pairsgeneration}

We generate all possible pairs of vertical and horizontal planes, regardless of the order.
Then, we compare all pairs in views $a$ and $b$ and estimate whether or not the two pairs $(i^a,j^a)$ and $(i^b,j^b)$ are composed of two plane matches $(i^a,i^b)$ and $(j^a,j^b)$.
A first test is performed using a simple penalty $e$ in plane pair space, that is agnostic to the view:

\begin{itemize}
    \item for pairs of non-parallel planes, i.e vertical non-parallel, the plane pair penalty is the difference of relative angles $\alpha$ between the normals of the planes:
    \begin{equation}
    \begin{array}{l}
    e_{(i^a,i^b),(j^a,j^b)} = | \alpha_{i^a j^a} - \alpha_{i^b j^b} | \\ = | \arccos(|\vec N_{i^a} . \vec N_{j^a}|) - \arccos(|\vec N_{i^b} . \vec N_{j^b}|) | \text{ ;}
    \end{array}
    \label{eq:matching_planepairdistnonpara}
    \end{equation}

    \item for pairs of parallel planes, i.e vertical parallel or horizontal, the plane pair penalty is the difference of relative distance $d$ in the normal directions of the planes:
    \begin{equation}
    \begin{array}{l}
    e_{(i^a,i^b),(j^a,j^b)} = | d_{i^a j^a} - d_{i^b j^b} | \\ = | \vec N_{i^a} . (P_{i^a} - P_{j^a}) - \vec N_{i^b} . (P_{i^b} - P_{j^b}) | \text{ .}
    \end{array}
    \label{eq:matching_planepairdistpara}
    \end{equation}
\end{itemize}

For each pair of planes in the two views, if this penalty $e$ is below a given threshold, we keep it and further check if the pair is an actual match, based on several heuristics, as explained below.
In practice, we keep a pair match if the difference is below 10 degrees in angle or 10 cm in distance.
These loose thresholds allow accounting for the noise disturbing the plane parameters between acquisitions.

\subsubsection{Plane Pairs Validation}
\label{sec:matching_association_pairsvalidtion}

First, we compute the motion matrix transforming the planes from one view to another, using the method from \autoref{sec:motion}.
For pairs of vertical planes, we compute the rotation (\autoref{sec:motion_rotation}) and horizontal translation (\autoref{sec:motion_transhoriz}).
For pairs of horizontal planes, we compute the vertical translation (\autoref{sec:motion_transvert}).
If the magnitude of the computed transformation is too high, we consider the match as wrong and discard it.
If it is low enough, we apply it to the second pair of planes and validate the match using plane-wise evaluation of:

\begin{itemize}
    \item the similarity of normal orientations and distances to origin;
    \item the overlap of convex hulls;
    \item the average Euclidean distance of 100 inliers of each plane to the other plane;
    \item the similarity of color histograms as described by Dou et al. \cite{dou2012exploring}.
\end{itemize}

After comparing all generated plane pairs, in case plane pair matches create more than one match for a single plane, we keep the match with the closest plane distance.
If there are not enough planes to generate pairs, or if no plane pairs have been matched, we consider single planes and compare them one by one using the previously described heuristics.
Finally, we recompute the motion between the two views using all plane matches derived from the pair matches, using the method described in \autoref{sec:motion}.

\section{Transformation Computation}
\label{sec:motion}

In this section, we describe a novel method to estimate the global transformation between two 3D views composed of matching planes,
based on the plane structural classification defined in \autoref{sec:matching_horizvert}.
We consider planes as horizontal and vertical and split the degrees of freedom of the transformation the same way.
In particular, while the rotation is computed in camera coordinate frame, the translation is first computed in world coordinate frame,
considering the known $Up$ vector as $Y$ axis, and then converted to camera coordinate frame.
Hence, the computation of the transformation is divided into three steps:

\begin{itemize}
	\item estimation of the rotation in camera space $X,Y,Z$;
	\item estimation of the horizontal translation with vertical planes along $X_w,Z_w$;
	\item estimation of the vertical translation with horizontal planes along $Y_w$.
\end{itemize}

As an optional pre-processing step, we refine the $Up$ vector in each view by taking the median value of deviations of horizontal plane normals to the current $Up$ vector.

\subsection{Rotation Computation}
\label{sec:motion_rotation}

\begin{figure*}
\centering
    \includegraphics[width=0.8\textwidth]{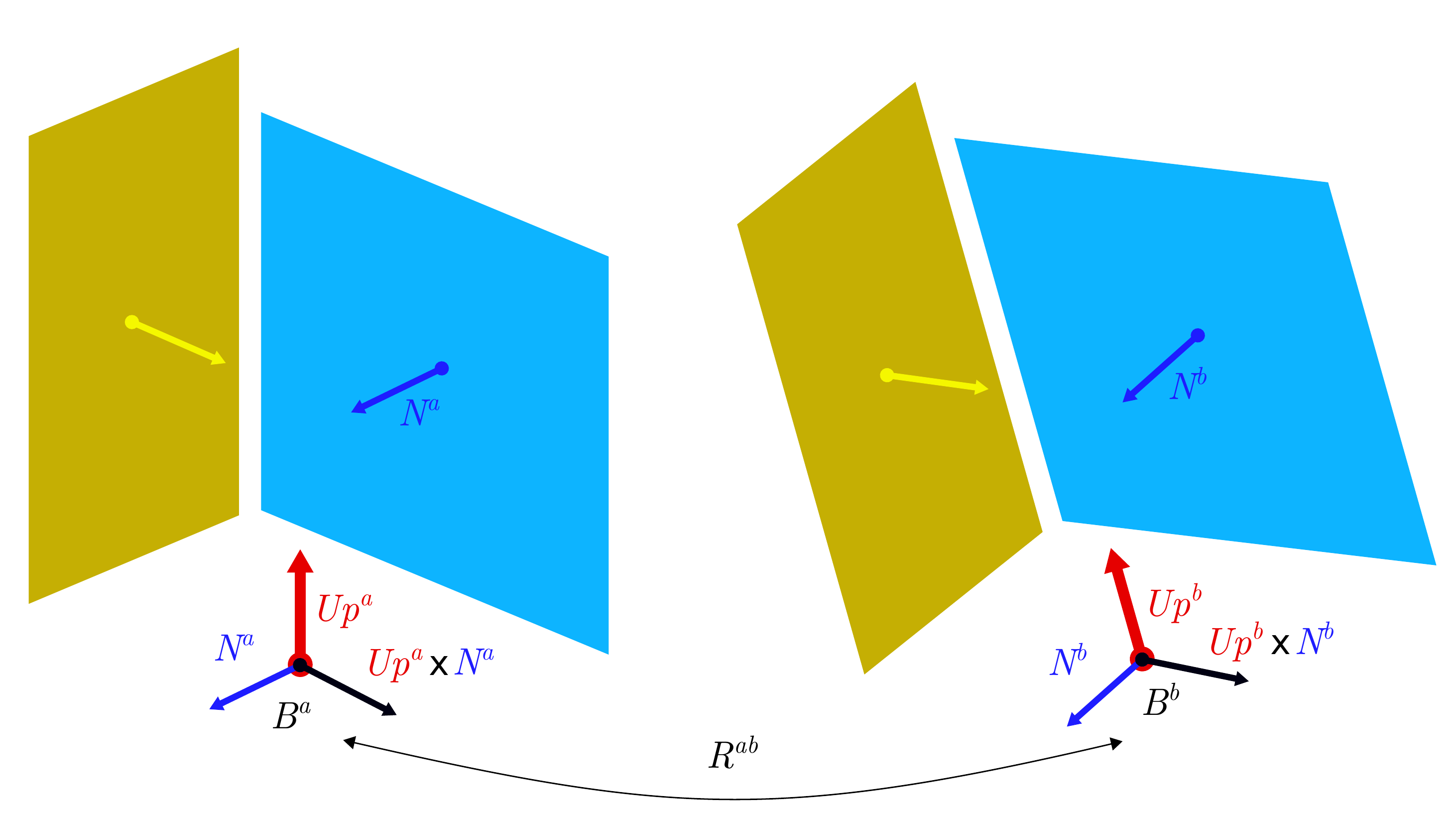}
\caption
{Rotation computation using a matching vertical plane.
As the $Up$ vector is known in both views, knowledge of a matching vertical plane of normal $N$ is enough to build a local coordinate frame $\{Up, ~ N, ~ Up \times N\}$.
Associating the local coordinate frames in both views $a$ and $b$ leads the complete rotation matrix between the views, as shown in \autoref{eq:motion_rotation}.
The same computation can be done for the second plane and rotations $R^{ab}$ can then be averaged.}
\label{fig:motion_rotation}
\end{figure*}

For views $a$ and $b$ of reference $Up$ vectors $Up^a$ and $Up^b$, the local coordinate frames for a plane of normal $N$ are given by $B^a$ and $B^b$.
The rotation matrix in world space can be inferred as $R^{ab}$, as shown in \autoref{eq:motion_rotation} and illustrated in \autoref{fig:motion_rotation}.
\begin{equation}
\begin{array}{l l l}
B^a &=& (N^a ~ Up^a ~ Up^a \times N^a) \in \mathbb{R}^{3 \times 3}\\
B^b &=& (N^b ~ Up^b ~ Up^b \times N^b) \in \mathbb{R}^{3 \times 3} \\
R^{ab} &=& B^{a ~ T} B^b \in \mathbb{R}^{3 \times 3} \\
\end{array}
\label{eq:motion_rotation}
\end{equation}

\subsection{Horizontal Translation}
\label{sec:motion_transhoriz}

We use the vertical planes matching between two views to compute the translation in the horizontal plane $X_w,Z_w$ of the world space.
First, we apply the rotation computed in the previous step and we project the plane normals and positions in this 2D space.
At this point, matching planes are aligned in orientation and are only transformed by a translation.

We have knowledge of the 2D normals and distances to origin of the planes in the two views, and the goal is to estimate the relative position of the camera origins in 2D horizontal space.
Given the relative orientations of the vertical planes, we have two choices:

\begin{itemize}
	\item if all planes are parallel and have the same direction, we can only compute the 2D translation in this direction;
	\item if at least one pair of planes is non-parallel, we use a \emph{quadric error minimizer} to compute the full translation in $X_w,Z_w$ world space.
\end{itemize}

\subsubsection{Parallel Planes}
\label{sec:motion_transhoriz_paral}

\begin{figure*}
\centering
    \includegraphics[width=0.8\textwidth]{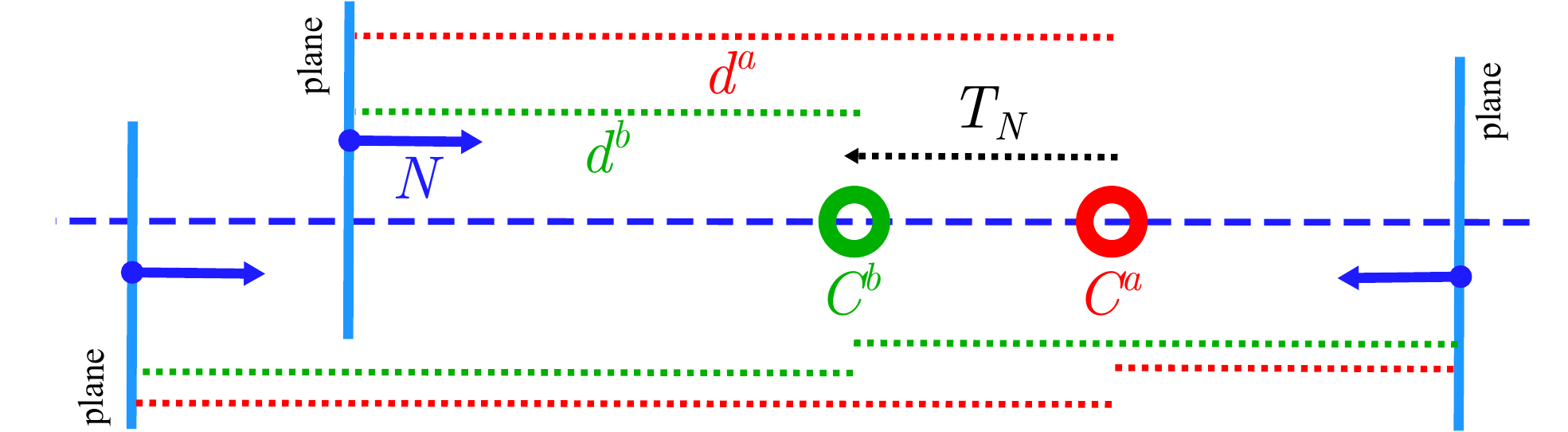}
\caption
{Computation of the translation in horizontal 2D $X_w,Z_w$ world space with planes (light blue) aligned along a single direction, here represented by the blue dotted line.
$C^a$ and $C^b$ are the centers of the camera in the two views, projected on this line.
The known distances to origin $d^a$ and $d^b$ (dotted red and green lines) of planes in the views allow recovering the value of the translation $T_N$ in the direction $N$ of the planes.}
\label{fig:motion_transonedir}
\end{figure*}

In the case where all vertical planes are parallel as defined in \autoref{sec:matching_horizvert},
we can only compute the translation of the camera along their normal direction in horizontal $X_w,Z_w$ world space, as illustrated in \autoref{fig:motion_transonedir}.
As we know the distances to origin of the plane in both views $d^a$ and $d^b$, we can compute their difference and apply it to the common direction to get a translation vector $T_N = (d^b - d^a) ~ N$.
This vector can then be converted into world space as $T_{X_w,Z_w} = [T_N.X_w ~ T_N.Z_w]$.

\subsubsection{Non-Parallel Planes}
\label{sec:motion_transhoriz_nonparal}

\begin{figure}[t]
\centering
    \includegraphics[width=\columnwidth]{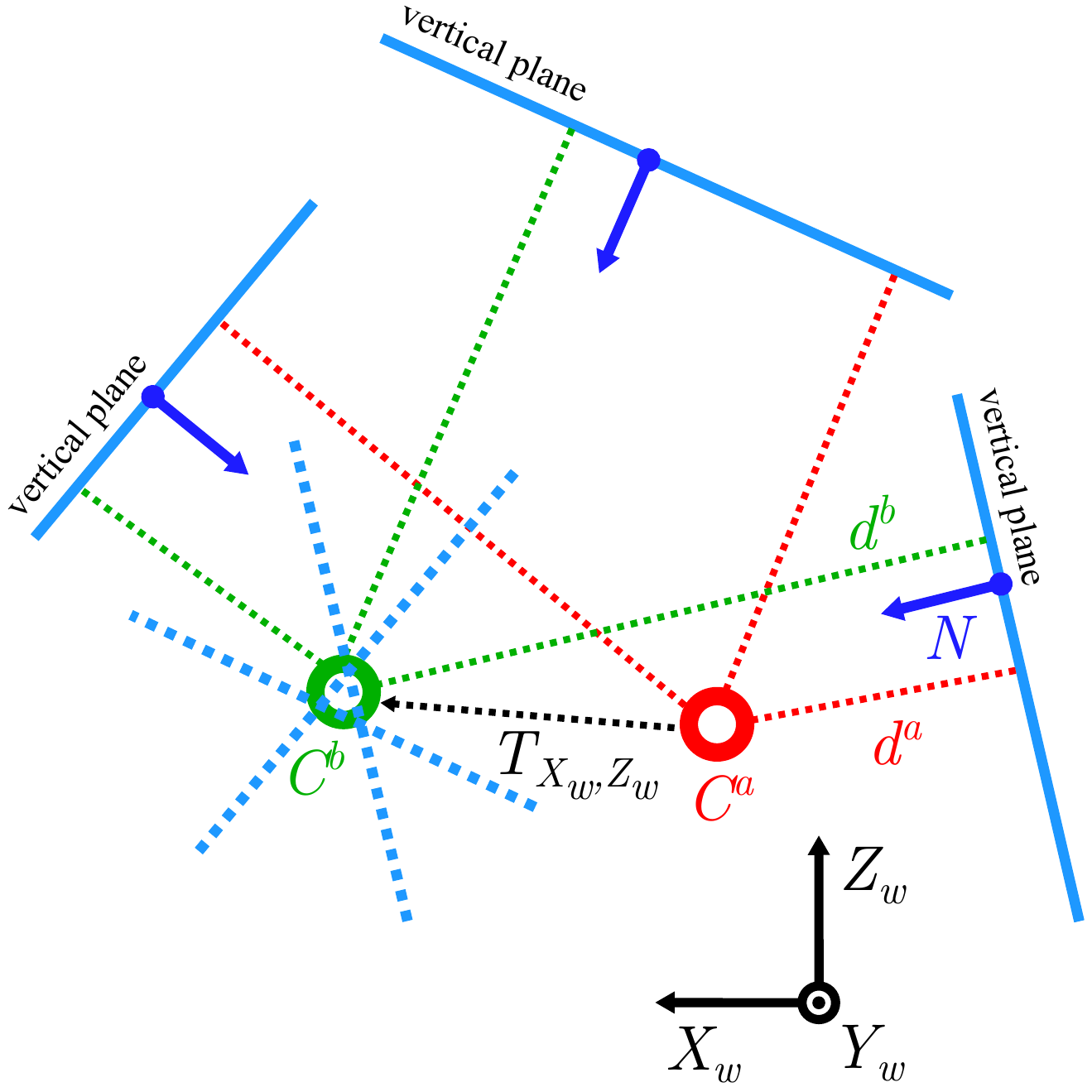}
\caption
{Computation of the translation in horizontal $X_w,Z_w$ world space with non-parallel plane directions.
$C^a$ and $C^b$ are the centers of the camera in the two views.
Blue lines with arrows represent tracked vertical planes projected in 2D $X_w,Z_w$ world space.
Dotted blue lines are the planes displaced of the distance to origin.
Dotted red and green lines represent the distances $d^a$ and $d^b$ of vertical planes to the centers of the camera in the first and second frames, respectively.
The quadric minimizer for those displaced planes leads the translation vector of the camera $T_{X_w,Z_w}$ in horizontal world space.}
\label{fig:motion_transhoriz_quadrics}
\end{figure}

In the case where there is at least one pair of non-parallel vertical planes, we can compute the full translation of the camera in horizontal $X_w,Z_w$ world space.
\autoref{fig:motion_transhoriz_quadrics} illustrates the setup in horizontal $X_w,Z_w$ world space where multiple planes have non-parallel directions.

In order to minimize the error, we make use of the quadric error to the planes and their minimizer point as described by Garland and Heckbert \cite{garland1997surface}.
For each vertical plane of normal $N$, we compute the fundamental error quadric in 2D space $X_w,Z_w$ as

\begin{equation}
\begin{array}{l l}
\multicolumn{2}{c}{K = pp^T}\\
&\\
\text{ with } p &= [a ~ b ~ -d]^T\\
&= [N.X_w ~ N.Z_w ~ -(d^a-d^b)]^T ~ .
\end{array}
\label{eq:motion_nonparallel}
\end{equation}

By defining the distance $d$ as the difference between distances to the origin of the planes in the first and second views,
we simulate the displacement of the planes to the origin of the second view, while taking the origin of the first as reference.
After summing the $K$ matrices for all planes, we solve for the minimizer as described by Garland and Heckbert, which describes the translation vector $T_{X_w,Z_w}$ along the $X_w$ and $Z_w$ axes in world space.

\subsection{Vertical Translation}
\label{sec:motion_transvert}

For the vertical translation, we consider the parallel case of the horizontal translation as illustrated in \autoref{fig:motion_transonedir},
with the Y world Up vector as common direction.
Using all horizontal planes, we compute the displacement along the Y axis as in \autoref{sec:motion_transhoriz_paral} with $N = Y_w$.

\section{Evaluation}
\label{sec:evaluation}

\definecolor{darkgreen}{rgb}{0.22,0.65,0.20}

In \autoref{sec:evaluation_toyexample}, we validate our motion computation method using a toy example.
Then, we evaluate both our plane matching (\autoref{sec:evaluation_matching}) and plane motion (\autoref{sec:evaluation_motion}) estimation strategies on recent public datasets presented in \autoref{sec:evaluation_dataset}.
We compare our methods to state-of-the-art 3D view pairwise registration methods.

\subsection{Toy Example}
\label{sec:evaluation_toyexample}

\begin{figure}
\centering
    \includegraphics[width=\columnwidth]{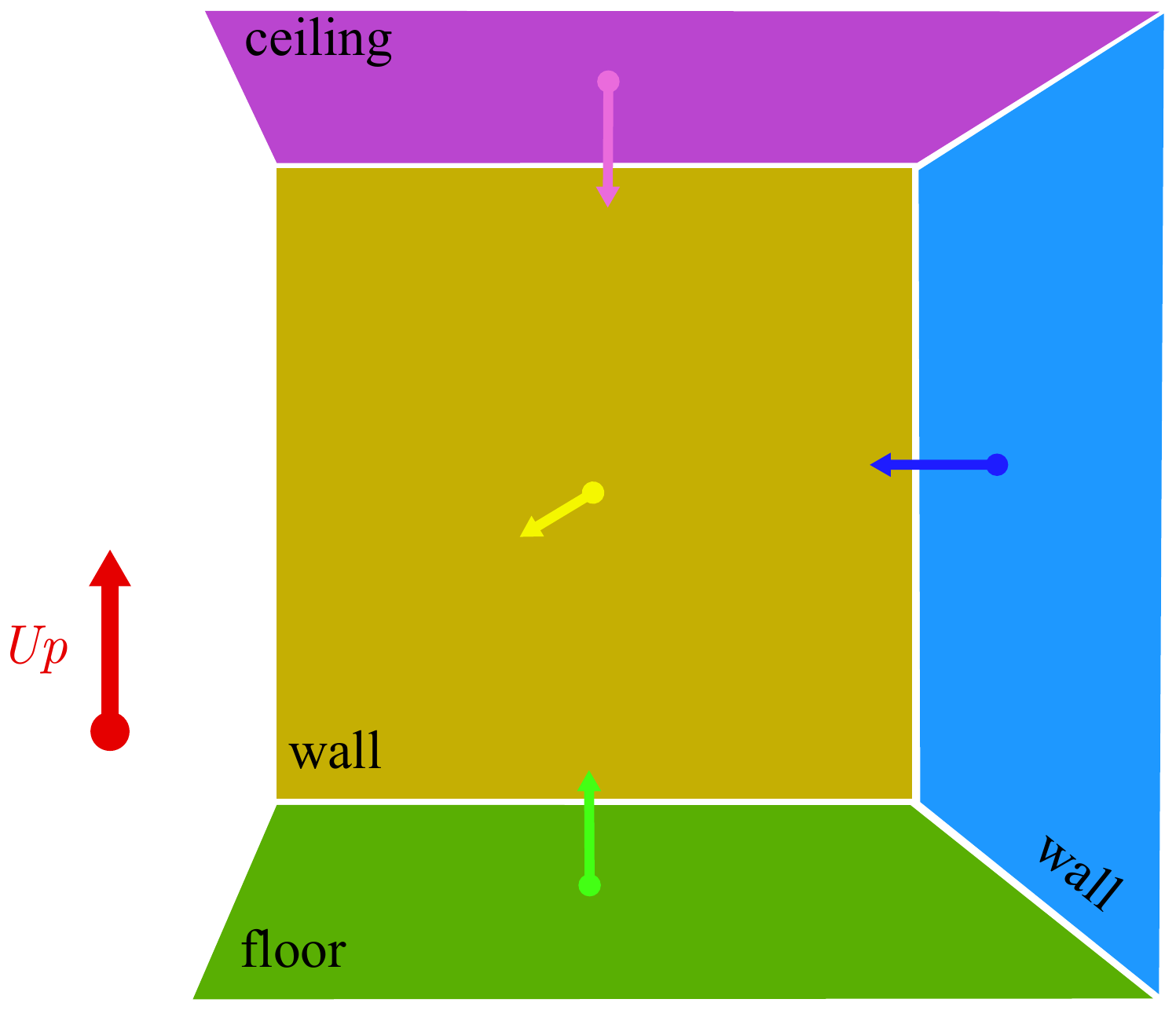}
\caption
{Toy example used to validate the plane-based transformation computation.
The red arrow is the $Up$ vector.
The blue and yellow planes are orthogonal vertical ones.
The green and purple are horizontal planes, modeling the floor and ceiling respectively.}
\label{fig:evaluation_toyexample}
\end{figure}

In order to validate the computation of the transformation as detailed in \autoref{sec:motion},
we designed a toy example, illustrated in \autoref{fig:evaluation_toyexample}.
The example models views of an indoor scene composed two orthogonal vertical planes, that can be seen as a left and far walls,
and two horizontal planes to represent the floor and ceiling.

With fixed initial plane parameters in one view, we randomly compute values of rotation of the camera with relation to the Up vector,
as well as a transformation matrix that we apply to the parameters, in order to run the registration algorithms on two views and recover the simulated motion.

\pgfplotstableread{
noiseperc roterrorrad roterrorradstd transerrormet transerrormetstd success valid
0   0.0    0.0    0.0    0.0    100    100
5   0.010  0.005  0.014  0.007  100    100
10  0.021  0.009  0.029  0.014  100    100
20  0.044  0.019  0.060  0.029  100    99.97
30  0.069  0.030  0.096  0.047  100    97.21
50  0.097  0.053  0.166  0.092  99.67  68.58
80  0.135  0.117  0.243  0.182  85.99  41.15
100 0.202  0.194  0.313  0.263  68.61  26.3
}
\registrationexperimentstoyexamplenoise

\begin{figure}[t!]
\centering
\begin{tikzpicture}
    \begin{axis}[
        width=0.9\columnwidth,
        height=8cm,
        axis y line=left, %
        axis x line=bottom, %
        axis line style={-}, %
        enlarge x limits=0.04,
        ylabel style={align=center, rotate=-90, at={(axis description cs:0.05,0.5)}},
        xlabel=Noise level (\%),
        ylabel=\textcolor{green!70!black}{\%},
        ymin=0,
        ymax=103]
        \addplot[smooth, mark=*, green!70!black] table [x=noiseperc, y=valid]{\registrationexperimentstoyexamplenoise}; \label{plotvalidity}
    \end{axis}
    \begin{axis}[
        width=0.9\columnwidth,
        height=8cm,
        axis y line=right, %
        axis x line=bottom, %
        axis line style={-}, %
        enlarge x limits=0.04,
        legend style={at={(0.5,1.25)},draw=none,anchor=north},
        legend cell align={left},
        ylabel style={align=center, rotate=-90, at={(axis description cs:0.98,0.5)}},
        xticklabels={,,},
        ylabel={\textcolor{blue!80!white}{rad} \\ or \\ \textcolor{red!80!white}{m}},
        ymin=0,
        ymax=0.63]
        \addlegendimage{/pgfplots/refstyle=plotvalidity}\addlegendentry{Validity (\%)}
        \addplot[smooth, mark=*, blue!80!white, error bars/.cd, y dir=both, y explicit] table [x=noiseperc, y=roterrorrad, y error=roterrorradstd]{\registrationexperimentstoyexamplenoise}; \addlegendentry{Rotation error (rad)}
        \addplot[smooth, mark=*, red!80!white, error bars/.cd, y dir=both, y explicit] table [x=noiseperc, y=transerrormet, y error=transerrormetstd]{\registrationexperimentstoyexamplenoise}; \addlegendentry{Translation error (m)}
    \end{axis}
\end{tikzpicture}
\caption
{Adding noise on the toy example for registration.
The rotation and translation errors are the norm along the three axes.
A registration is considered valid if its rotation angle error is below 20 degrees and its translation error is below 20cm.
Each noise level experiment was ran 10000 times and averaged.
This experiment first validates our motion estimation in the perfect case where no noise is present.
Then, it shows a good robustness to noisy point sets up to about 30\% of noise. %
The use of planes even in noisy depth maps allows smoothing the error and recovering correct information in many cases.}
\label{fig:evaluation_toyexample_noiseplot}
\end{figure}
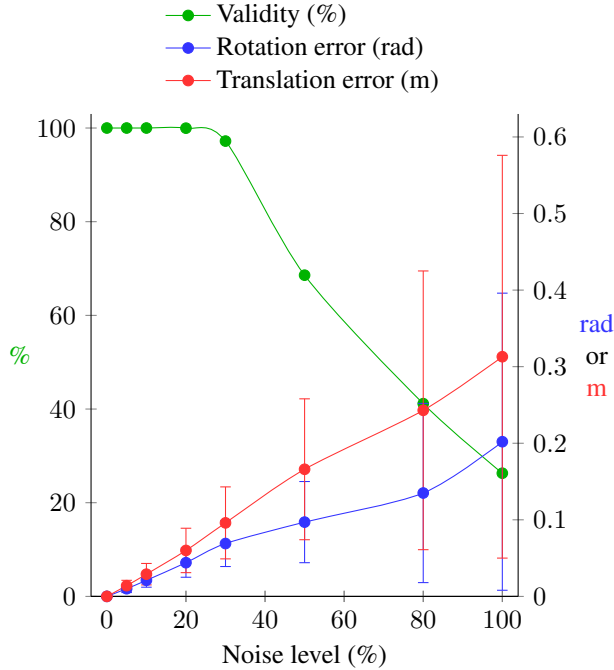

We show experiments on this toy example while adding uniform noise in \autoref{fig:evaluation_toyexample_noiseplot}.
To simulate sensor noise, we sample the plane surface in both views with 400 random 3D point locations after applying the random transformation to the plane parameters.
We then apply uniform noise displacement to the points as a percentage of a maximum value of 2m, which ensures covering most noise and outlier values observed in practice.
Finally, we re-estimate the normals and centers of the planes using \emph{singular value decomposition} of the noisy inlier set
and use them in the plane-based registration algorithm.
Although this noise model is not the one observed on samples from depth sensors,
it gives a first insight on the influence of noise on a controllable example, while staying simple and general.
In addition, while we only generate 400 noisy points, in practice with real captured data
the aggregation of thousands of inlier point positions to compute the plane parameters allows the noise in the observed data to stay low.

In \autoref{fig:evaluation_toyexample_noiseplot}, we can see that with no noise added to the plane inliers, the transformation matrix is always recovered with no error,
which validates our motion estimation error.
Although, we notice that the accuracy of the transformation quickly drops after 30\% of added synthetic noise,
which highlights the sensitivity of our algorithm to noisy planes.

\subsection{Dataset}
\label{sec:evaluation_dataset}

In order to evaluate the accuracy of our plane-based registration, we make use of the publicly available RGB-D dataset \emph{SUN3D} \cite{xiao2013sun3d}.
In particular, we use the benchmark provided by the authors of \emph{fine-to-coarse} \cite{halber2017finetocoarse} and \emph{3DMatch} \cite{zeng20163dmatch}.
The former provides ground truth correspondences in overlapping RGB-D frames, which we use to compute error values and evaluate the accuracy of our registration.
The latter, similar to the synthetic benchmark of Choi et al. \cite{choi2015robust}, provides scene fragments generated from 50 fused RGB-D frames at 6mm resolution, associated with ground truth transformations and correspondences.

As specified by Choi et al., we consider a registered pair as valid if the average error of all correspondences
after applying the transformation, computed as the averaged Euclidean distance between transformed corresponding 3D point locations, falls below the threshold of \emph{20cm}.
We then attempt to register pairs of frames and fragments with available ground truth motion and correspondences.

We quantitatively evaluate the registration with the following metrics:

\begin{itemize}
    \item \emph{success}: the amount of pairs that the algorithm is able to register;
    \item \emph{recall}: the amount of correct pairs that the algorithm successfully registers and are valid;
    \item \emph{precision}: the amount of registered pairs that are valid, i.e. with a correspondence error below the threshold;
    \item \emph{root mean squared error (RMSE)}: the average error of the ground truth correspondences for all registered pairs;
    \item \emph{median absolute error (MAE)}: the median error of the ground truth correspondences for all registered pairs;
    \item \emph{time}: the average total number of milliseconds required to register a pair.
\end{itemize}

We compare our plane-based registration to state-of-the-art \emph{pairwise} registration methods using \emph{ORB} keypoint descriptors in RGB image \cite{rublee2011orb} and \emph{fast point feature histograms (FPFH)} \cite{rusu2009fast}.
In the following tables, we present the evaluation metrics of our methods with relative difference to the accuracy of \emph{ORB} and \emph{FPFH}.

\subsection{Plane Matching}
\label{sec:evaluation_matching}

To evaluate our plane matching strategy presented in \autoref{sec:matching}, we run plane detection in both frames separately
and feed the planes to the algorithm to match them and estimate the motion between the views.
The method will be successful as long as the frames have a common vertical plane that is correctly identified.
However, the validity of the estimated motion matrix -- or precision -- is evaluated using the ground truth motion and correspondences provided by the datasets.

\autoref{tab:evaluation_rgbd} shows quantitative evaluation of single RGB-D frames registration using our plane matching method,
while \autoref{tab:evaluation_fragments} shows evaluation of scan fragments registration.
In both cases, while the plane-based method shows significant speed-up over \emph{FPFH}, we can see a noticeable degradation of the accuracy.
We put that degradation down to the lack of distinguishable features between geometrically similar planes, even when taking appearance histograms into account.
In consequence, we consider a prior registration estimate, however coarse, as needed to match the planes before computing the transformation.

\subsection{Motion Computation}
\label{sec:evaluation_motion}

As we notice that geometric planes are not characteristic enough by themselves to be matched even when classified and grouped by pairs,
we first compute a coarse transformation matrix to track the planes as presented in \autoref{sec:matching_tracking},
and then refine this estimate using our method presented in \autoref{sec:motion}.
The method will be successful as long as the prior registration is successful and the frames in the pair have a common horizontal or vertical plane.
In the case where some planes are missing to compute the full 6 degrees of freedom of the transformation,
we use the prior estimate to fill in for the components of the rotation and translation that we can not compute.

Again, we compute evaluation metrics for the registration of single RGB-D frames in \autoref{tab:evaluation_rgbd}
and scan fragments in \autoref{tab:evaluation_fragments}.
For RGB-D frames, computing a prior transformation with \emph{ORB} keypoints leads to a 10 to 20\% increase in precision, while reducing the median error of about 10\%.
The significant speed-up of 90\% over \emph{FPFH} shows a real interest of the plane-based method when coupled with \emph{ORB}.
However, using \emph{FPFH} as a prior registration method, while allowing more frame pairs to succeed, reduces the accuracy of the transformation.
For scan fragments, the plane-based method does not improve accuracy, as the prior \emph{FPFH} motion estimate is likely accurate enough.
As scan fragments aggregate the information from 50 single frames, more distinctive geometric features should overlap between the views, allowing the \emph{FPFH} descriptor to perform well.

\begin{table*}
\setlength{\tabcolsep}{12pt} %
\centering
\begin{tabular}{l c c c c c c}
    \hline
    Method & Success (\%) & Rec (\%) & Prec (\%) & RMSE (m) & MAE (m) & Time (ms) \\
    \hline
    ORB & 78.1 & 42.7 & 53.2 & 0.771 & 0.133 & 8 \\
    FPFH & 73.5 & 41.8 & 55.6 & 0.649 & 0.141 & 2470 \\
    \hline
    Planes only & 38.7 & 11.4 & 27.1 & 1.01 & 0.483 & 396 \\
    \multicolumn{1}{r}{ \footnotesize vs ORB }  & \textcolor{red}{\footnotesize -50\%} & \textcolor{red}{\footnotesize -73\%} & \textcolor{red}{\footnotesize -49\%} & \textcolor{red}{\footnotesize +31\%} & \textcolor{red}{\footnotesize +263\%} & \textcolor{red}{\footnotesize +5K\%} \\
    \multicolumn{1}{r}{ \footnotesize vs FPFH } & \textcolor{red}{\footnotesize -47\%} & \textcolor{red}{\footnotesize -73\%} & \textcolor{red}{\footnotesize -51\%} & \textcolor{red}{\footnotesize +56\%} & \textcolor{red}{\footnotesize +243\%} & \textcolor{darkgreen}{\footnotesize -84\%} \\
    \hline
    ORB + planes & 52.4 & 34.4 & 63.3 & 0.732 & 0.122 & 228 \\
    \multicolumn{1}{r}{ \footnotesize vs ORB }  & \textcolor{red}{\footnotesize -33\%} & \textcolor{red}{\footnotesize -19\%} & \textcolor{darkgreen}{\footnotesize +19\%} & \textcolor{darkgreen}{\footnotesize -5\%} & \textcolor{darkgreen}{\footnotesize -8\%} & \textcolor{red}{\footnotesize +3K\%} \\
    \rowcolor{black!10!white}
    \multicolumn{1}{r}{ \footnotesize \textbf{vs FPFH} } & \textcolor{red}{\footnotesize -29\%} & \textcolor{red}{\footnotesize -18\%} & \textcolor{darkgreen}{\footnotesize +14\%} & \textcolor{red}{\footnotesize +13\%} & \textcolor{darkgreen}{\footnotesize -13\%} & \textcolor{darkgreen}{\footnotesize -91\%} \\
    FPFH + planes & 63.7 & 33.6 & 52.0 & 0.837 & 0.169 & 2668 \\
    \multicolumn{1}{r}{ \footnotesize vs ORB }  & \textcolor{red}{\footnotesize -18\%} & \textcolor{red}{\footnotesize -21\%} & \textcolor{red}{\footnotesize -2\%} & \textcolor{red}{\footnotesize +9\%} & \textcolor{red}{\footnotesize +27\%} & \textcolor{red}{\footnotesize +33K\%} \\
    \multicolumn{1}{r}{ \footnotesize vs FPFH } & \textcolor{red}{\footnotesize -13\%} & \textcolor{red}{\footnotesize -20\%} & \textcolor{red}{\footnotesize -6\%} & \textcolor{red}{\footnotesize +29\%} & \textcolor{red}{\footnotesize +20\%} & \textcolor{red}{\footnotesize +8\%} \\
    \hline
\end{tabular}
\caption
{Registration of single RGB-D frames using our plane matching strategy as well as planes tracked with a prior estimate.
Using \emph{ORB} as prior allows improving accuracy, while showing significant speed-up over \emph{FPFH} (highlighted row).}
\label{tab:evaluation_rgbd}
\end{table*}

\begin{table*}
\setlength{\tabcolsep}{12pt} %
\centering
\begin{tabular}{l c c c c c c}
    \hline
    Method & Success (\%) & Rec (\%) & Prec (\%) & RMSE (m) & MAE (m) & Time (ms) \\
    \hline
    FPFH & 90.2 & 54.1 & 68.2 & 1.13 & 0.100 & 3004 \\
    \hline
    Planes only & 73.7 & 14.3 & 21.0 & 1.254 & 0.865 & 987 \\
    \multicolumn{1}{r}{ \footnotesize vs FPFH } & \textcolor{red}{\footnotesize -18\%} & \textcolor{red}{\footnotesize -74\%} & \textcolor{red}{\footnotesize -69\%} & \textcolor{red}{\footnotesize +11\%} & \textcolor{red}{\footnotesize +765\%} & \textcolor{darkgreen}{\footnotesize -67\%} \\
    \hline
    FPFH + planes & 90.1 & 28.3 & 35.8 & 1.26 & 0.365 & 3855 \\
    \multicolumn{1}{r}{ \footnotesize vs FPFH } & \textcolor{red}{\footnotesize -0\%} & \textcolor{red}{\footnotesize -48\%} & \textcolor{red}{\footnotesize -48\%} & \textcolor{red}{\footnotesize +12\%} & \textcolor{red}{\footnotesize +265\%} & \textcolor{red}{\footnotesize +28\%} \\
    \hline
\end{tabular}
\caption
{Registration of scan fragments using our plane matching strategy as well as planes tracked with a prior estimate.
The higher amount of information present in fragments allows \emph{FPFH} to perform better than the plane-based method.}
\label{tab:evaluation_fragments}
\end{table*}

\begin{figure}[t]
\centering
    \includegraphics[width=\columnwidth]{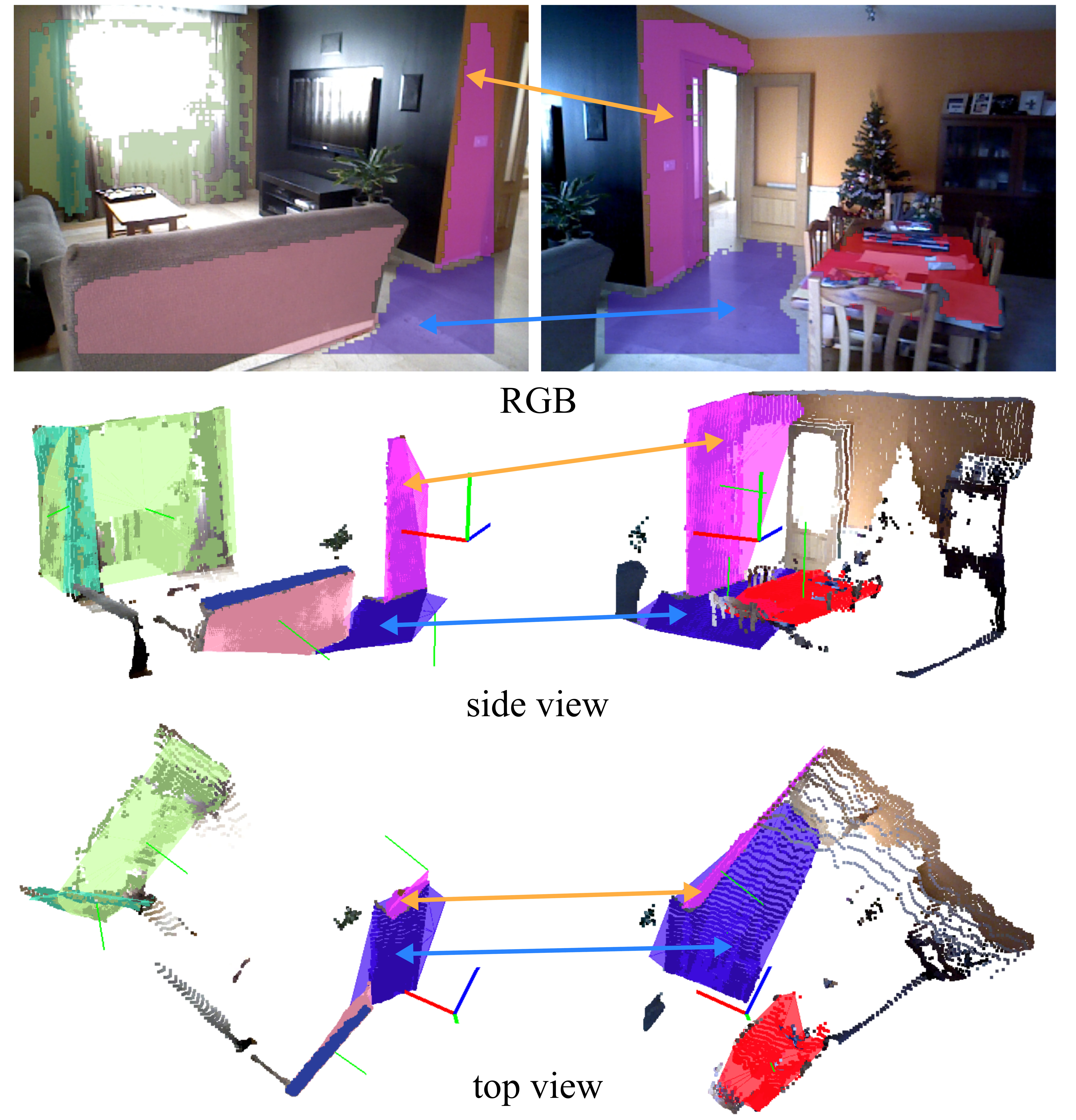}
\caption
{Corresponding planes in two RGB-D views used to compute their relative transformation.
In this example, we use frames 301 and 401 of scene \emph{home\_at/home\_at\_scan1\_2013\_jan\_1} from the \emph{SUN3D} database.
Top: plane inliers overlaid on RGB images.
Bottom: side and top views showing detected planes in 3D with their normal vectors in thin green and the world frame in red, green and blue lines.
Only the blue horizontal (blue arrows) and purple vertical (orange arrows) planes are common in the two views, but they are sufficient to compute the sensor motion.}
\label{fig:evaluation_framesplanes}
\end{figure}

\begin{figure*}
\centering
    \includegraphics[width=0.8\textwidth]{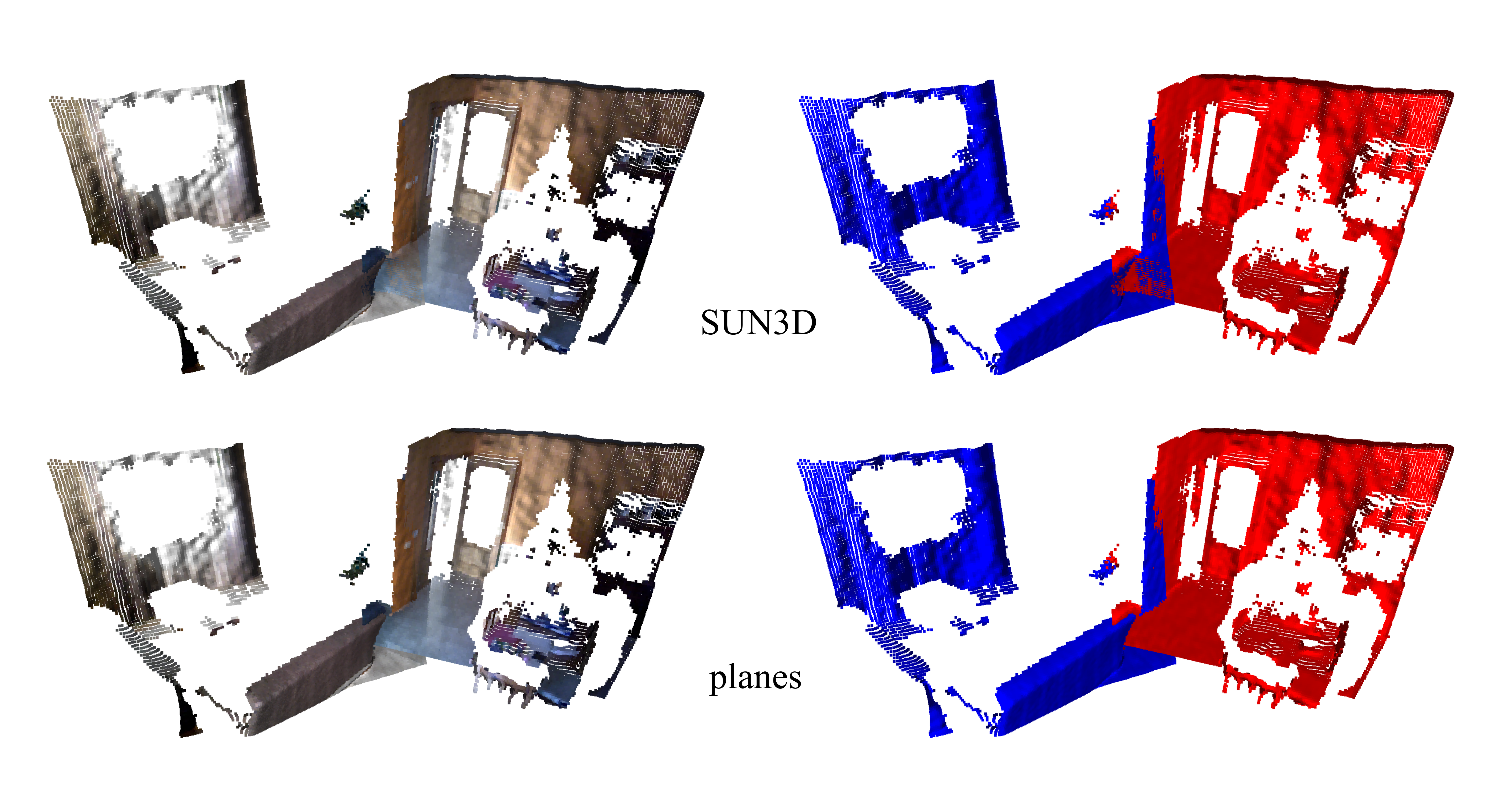}
\caption
{Registered RGB-D views in a common space using \emph{SUN3D SfM} \cite{xiao2013sun3d},
giving matrices as ground truth in the dataset (top), and our plane-based method (bottom).
In these two frames, the overlap is rather low and the overlapping region does not contain many distinctive features,
which are essential to the performance of regular 3D descriptors.
Here, recovering two overlapping planes allows us to compute the full transformation between sensor positions.
We can see that the difference with the ground truth is subtle, showing the high quality of the plane-based registration.}
\label{fig:evaluation_framessamespace}
\end{figure*}

\autoref{fig:evaluation_framesplanes} shows corresponding planes in two views used to compute the transformation.
\autoref{fig:evaluation_framessamespace} shows the two registered views in a common space.
In this example, we can see that even low overlap is sufficient to register the views if enough common planes are present.
Here in both frames, the floor plane allows recovering the gravity vector and its association with the wall planes leads the full three angles rotation.
Translation is then recovered using the floor (vertical translation) and wall (horizontal translation).

\section{Conclusion} %
\label{sec:conclusion}

In this paper, we introduced planar geometric and structural constraints into the problem of 3D view registration.
We have presented multiple ways to make use of planar structures seen in overlapping 3D views in order to estimate the motion between them.
We exploit the \emph{Manhattan world} structure of indoor scenes to recover the motion of the sensor with relation to this structure.
This allows the methods to be lightweight and makes them agnostic to subtle changes in the positions of objects.
While results could be improved, we like to see these contributions as incremental and designed to be paired with other existing methods in order to reach state-of-the-art accuracy and performance.
We discuss them below and give hints towards improvement of robustness and accuracy.

\subsection{Discussion and Limitations}
\label{sec:conclusion_discussion}

The analysis of the scene structure, agnostic to subtle changes in scene geometry of e.g., small objects,
is particularly well suited to lifelong 3D information aggregation.
In that sense, our use of large planar structures to register views with each other is an advantage over state-of-the-art local descriptors.
In particular, the analysis of relations between these structures and the definition of plane pair distances that are agnostic to the point of view adds robustness to the registration process.

Computing the sensor motion from separated horizontal and vertical planes to split computation of rotations and translations
works well in practice when used as a refinement step after a prior coarse estimation method.
We show improvement in accuracy over the prior method, as well as fast processing compared to 3D descriptors.
However, the requirement for overlapping planar structures in the two views leads to a lower success rate.

For view matching, detected planes do not seem to be strong distinguishable features in an indoor scene, where more complex 3D descriptors give better results.
The grouping of planes by pairs, while reducing complexity of the problem, does not prevent wrong plane matches as much as expected.
On the other hand, the low success rate can be explained by the requirement for overlapping \emph{Manhattan world} structures in the views, which are not always present.

\subsection{Towards Robust Shape-based Registration} %
\label{sec:conclusion_challenges}

While geometric planes offer a more stable support than 3D points to estimate the motion of a sensor in a scene,
they also embed less information that regular 3D descriptors and can lead to some confusion.
We give here multiple research paths and challenges for the improvement of robustness in plane distinction and estimation of the transformation matrix,
in order to improve current results.

\subsubsection{Plane Matching}

In order to ease the distinction of different geometrically similar planes and prevent confusion, we could use stronger heuristics.
When color information is available, the comparison of color keypoints detected in the rectified 2D space of the plane could help discarding wrong matches.

While we presented associations of two planes to form pairs and their relative distances, we could imagine grouping planes by three or four.
This would require designing specific descriptors based on geometric information in order to discriminate the correct groups from each other.

Finally, we could add a quality check to discard wrong plane pair matches.
By applying the estimated transformation for a plane pair match to other already detected plane pair matches,
we could identify whether the motion matrix correctly transforms planes from a view to another,
and discard the new plane pair if not.

\subsubsection{Transformation Computation}

In order to reduce error on the translation, when possible, we could use an horizontal plane associated with two vertical non-parallel planes.
After applying the rotation, their unique intersection point will allow computing the full translation of the sensor.

We could imagine lowering the quality of the prior 3D descriptor method to speed it up, while keeping enough accuracy for plane tracking.
This would allow faster processing with a plane-based refinement step to reach state-of-the-art accuracy.

\subsubsection{Extension to Objects}

A limitation of our approach is the need for overlapping planar structures in two views, which is not always the case.
To increase the success rate of our method, we could use not only planes, but also objects present in the scene, that would be modeled by simple shapes as well.
By exploiting the parameters of simple shapes such as cylinders, spheres or boxes matched in overlapping views, in the same spirit as with planes, we could infer sensor motion more accurately.

\subsubsection{Evaluation}

In order to further understand the behavior of both algorithms in failure cases, we could investigate specific cases by implementing visual quality check.
In addition, the model of the noise we add to the toy example is rather far from the noise observed in captured data.
We could imagine using a more advanced depth sensor simulator on our synthetic dataset in order to get a more meaningful evaluation.

{\small
\bibliographystyle{abbrv}
\bibliography{content/main}

\begin{thebibliography}{10}

\bibitem{aiger20084points}
D.~Aiger, N.~J. Mitra, and D.~Cohen-Or.
\newblock 4-points congruent sets for robust pairwise surface registration.
\newblock {\em ACM Transactions on Graphics (TOG)}, 27(3):85, 2008.

\bibitem{bay2006surf}
H.~Bay, T.~Tuytelaars, and L.~Van~Gool.
\newblock {SURF: Speeded up robust features}.
\newblock In {\em European Conference on Computer Vision}, pages 404--417.
  Springer, 2006.

\bibitem{bellekens2014survey}
B.~Bellekens, V.~Spruyt, R.~Berkvens, and M.~Weyn.
\newblock {A Survey of Rigid 3D Pointcloud Registration Algorithms}.
\newblock In {\em AMBIENT 2014: the 4th International Conference on Ambient
  Computing, Applications, Services and Technologies, August 24-28, 2014, Rome,
  Italy}, pages 8--13, August 2014.

\bibitem{besl1992method}
P.~J. Besl and N.~D. McKay.
\newblock {A Method for Registration of 3-D Shapes}.
\newblock {\em IEEE Transactions on Pattern Analysis and Machine Intelligence},
  14(2):239--256, 1992.

\bibitem{bouaziz2013sparseicp}
S.~Bouaziz, A.~Tagliasacchi, and M.~Pauly.
\newblock {Sparse Iterative Closest Point}.
\newblock {\em Computer Graphics Forum (Symposium on Geometry Processing)},
  32(5):1--11, 2013.

\bibitem{chen1991object}
Y.~Chen and G.~Medioni.
\newblock {Object Modelling by Registration of Multiple Range Images}.
\newblock In {\em Proceedings. 1991 IEEE International Conference on Robotics
  and Automation}, pages 2724--2729. Elsevier, 1991.

\bibitem{coughlan1999manhattan}
J.~M. Coughlan and A.~L. Yuille.
\newblock Manhattan world: Compass direction from a single image by bayesian
  inference.
\newblock In {\em Proceedings of the 7th IEEE International Conference on
  Computer Vision}, volume~2, pages 941--947. IEEE, 1999.

\bibitem{deng2018ppfnet}
H.~Deng, T.~Birdal, and S.~Ilic.
\newblock {PPFNet: Global Context Aware Local Features for Robust 3D Point
  Matching}.
\newblock In {\em Computer Vision and Pattern Recognition (CVPR), 2018 IEEE
  Conference on}, 2018.

\bibitem{dou2012exploring}
M.~Dou, L.~Guan, J.-M. Frahm, and H.~Fuchs.
\newblock {Exploring High-Level Plane Primitives for Indoor 3D Reconstruction
  with a Hand-held RGB-D Camera}.
\newblock In {\em Asian Conference on Computer Vision}, pages 94--108.
  Springer, 2012.

\bibitem{drost2010model}
B.~Drost, M.~Ulrich, N.~Navab, and S.~Ilic.
\newblock {Model Globally, Match Locally: Efficient and Robust 3D Object
  Recognition}.
\newblock In {\em Computer Vision and Pattern Recognition (CVPR), 2010 IEEE
  Conference on}, pages 998--1005. Ieee, 2010.

\bibitem{endres20143dmapping}
F.~Endres, J.~Hess, J.~Sturm, D.~Cremers, and W.~Burgard.
\newblock {3D Mapping with an RGB-D Camera}.
\newblock {\em IEEE Transactions on Robotics}, 30(1):177--187, 2014.

\bibitem{fischler1981random}
M.~A. Fischler and R.~C. Bolles.
\newblock {Random Sample Consensus: A Paradigm for Model Fitting with
  Applications to Image Analysis and Automated Cartography}.
\newblock {\em Communications of the ACM}, 24(6):381--395, June 1981.

\bibitem{forstner2017efficient}
W.~F{\"o}rstner and K.~Khoshelham.
\newblock {Efficient and Accurate Registration of Point Clouds with Plane to
  Plane Correspondences}.
\newblock In {\em Computer Vision Workshop, IEEE International Conference on},
  pages 2165--2173. IEEE, 2017.

\bibitem{garland1997surface}
M.~Garland and P.~S. Heckbert.
\newblock {Surface Simplification using Quadric Error Metrics}.
\newblock In {\em Proceedings of the 24th annual conference on Computer
  graphics and interactive techniques}, pages 209--216. ACM
  Press/Addison-Wesley Publishing Co., August 1997.

\bibitem{guo2016comprehensive}
Y.~Guo, M.~Bennamoun, F.~Sohel, M.~Lu, J.~Wan, and N.~M. Kwok.
\newblock {A Comprehensive Performance Evaluation of 3D Local Feature
  Descriptors}.
\newblock {\em International Journal of Computer Vision}, 116(1):66--89,
  January 2016.

\bibitem{halber2017finetocoarse}
M.~Halber and T.~Funkhouser.
\newblock {Fine-To-Coarse Global Registration of RGB-D Scans}.
\newblock In {\em IEEE Conference on Computer Vision and Pattern Recognition
  (CVPR)}, 2017.

\bibitem{johnson1999using}
A.~E. Johnson and M.~Hebert.
\newblock {Using Spin Images for Efficient Object Recognition in Cluttered 3D
  Scenes}.
\newblock {\em IEEE Transactions on Pattern Analysis \& Machine Intelligence},
  21(5):433--449, 1999.

\bibitem{kaiser2018survey}
A.~Kaiser, J.~A. Ybanez~Zepeda, and T.~Boubekeur.
\newblock {A Survey of Simple Geometric Primitives Detection Methods for
  Captured 3D Data}.
\newblock {\em Computer Graphics Forum}, 38(1):167--196, February 2019.

\bibitem{lowe1999sift}
D.~G. Lowe.
\newblock {Object Recognition from Local Scale-invariant Features}.
\newblock {\em International Conference on Computer Vision}, 99(2):1150--1157,
  1999.

\bibitem{mellado2014super4pcs}
N.~Mellado, D.~Aiger, and N.~J. Mitra.
\newblock {Super4PCS: Fast Global Pointcloud Registration via Smart Indexing}.
\newblock {\em Computer Graphics Forum}, 33(5):205--215, 2014.

\bibitem{morell2018survey}
V.~Morell-Gimenez, M.~Saval-Calvo, V.~Villena-Martinez, J.~Azorin-Lopez,
  J.~Garcia-Rodriguez, M.~Cazorla, S.~Orts-Escolano, and A.~Fuster-Guillo.
\newblock {A survey of 3D rigid registration methods for RGB-D cameras}.
\newblock {\em Advancements in Computer Vision and Image Processing}, pages
  74--98, 2018.

\bibitem{newcombe2011kinectfusion}
R.~A. Newcombe, S.~Izadi, O.~Hilliges, D.~Molyneaux, D.~Kim, A.~J. Davison,
  P.~Kohi, J.~Shotton, S.~Hodges, and A.~Fitzgibbon.
\newblock {KinectFusion: Real-time Dense Surface Mapping and Tracking}.
\newblock In {\em IEEE International Symposium on Mixed and Augmented Reality
  (ISMAR)}, pages 127--136. IEEE, October 2011.

\bibitem{pathak2010fast}
K.~Pathak, A.~Birk, N.~Vaskevicius, and J.~Poppinga.
\newblock {Fast Registration Based on Noisy Planes with Unknown Correspondences
  for 3D Mapping}.
\newblock {\em IEEE Transactions on Robotics}, 26(3):424--441, 2010.

\bibitem{rublee2011orb}
E.~Rublee, V.~Rabaud, K.~Konolige, and G.~Bradski.
\newblock {ORB: An efficient alternative to SIFT or SURF}.
\newblock In {\em International Conference on Computer Vision}, pages
  2564--2571. IEEE, 2011.

\bibitem{rusinkiewicz2001efficient}
S.~Rusinkiewicz and M.~Levoy.
\newblock {Efficient Variants of the ICP Algorithm}.
\newblock {\em 3DIM}, pages 145--152, May 2001.

\bibitem{rusu2009fast}
R.~B. Rusu, N.~Blodow, and M.~Beetz.
\newblock {Fast Point Feature Histograms (FPFH) for 3D Registration}.
\newblock In {\em Robotics and Automation, 2009. ICRA'09. IEEE International
  Conference on}, pages 3212--3217. IEEE, 2009.

\bibitem{rusu2008aligning}
R.~B. Rusu, N.~Blodow, Z.~C. Marton, and M.~Beetz.
\newblock {Aligning Point Cloud Views using Persistent Feature Histograms}.
\newblock In {\em 2008 IEEE/RSJ International Conference on Intelligent Robots
  and Systems}, pages 3384--3391. IEEE, 2008.

\bibitem{schnabel2007efficient}
R.~Schnabel, R.~Wahl, and R.~Klein.
\newblock {Efficient RANSAC for Point-Cloud Shape Detection}.
\newblock {\em Computer Graphics Forum}, 26(2):214--226, June 2007.

\bibitem{segal2009generalized}
A.~Segal, D.~Haehnel, and S.~Thrun.
\newblock {Generalized-ICP}.
\newblock {\em Robotics Science and Systems}, 5, June 2009.

\bibitem{shi2018planematch}
Y.~Shi, K.~Xu, M.~Nie{\ss}ner, S.~Rusinkiewicz, and T.~Funkhouser.
\newblock {PlaneMatch: Patch Coplanarity Prediction for Robust RGB-D
  Reconstruction}.
\newblock In {\em Proceedings of the European Conference on Computer Vision
  ({ECCV})}, 2018.

\bibitem{choi2015robust}
{Sungjoon Choi and Qian-Yi Zhou and Vladlen Koltun}.
\newblock Robust reconstruction of indoor scenes.
\newblock In {\em IEEE Conference on Computer Vision and Pattern Recognition},
  2015.

\bibitem{umeyama1991least}
S.~Umeyama.
\newblock Least-squares estimation of transformation parameters between two
  point patterns.
\newblock {\em IEEE Transactions on pattern analysis and machine intelligence},
  13(4):376--380, 1991.

\bibitem{viejo20083dmodel}
D.~Viejo~Hernando and M.~Cazorla.
\newblock {3D Model Based Map Building}.
\newblock In {\em IX Workshop de Agentes Fisicos (WAF'2008), Vigo, Spain},
  September 2008.

\bibitem{xiao2013sun3d}
J.~Xiao, A.~Owens, and A.~Torralba.
\newblock {SUN3D: A Database of Big Spaces Reconstructed using SfM and Object
  Labels}.
\newblock In {\em Proceedings of the IEEE International Conference on Computer
  Vision}, pages 1625--1632, 2013.

\bibitem{zeng20163dmatch}
A.~Zeng, S.~Song, M.~Nie{\ss}ner, M.~Fisher, J.~Xiao, and T.~Funkhouser.
\newblock {3DMatch: learning local geometric descriptors from RGB-D
  reconstructions}.
\newblock In {\em CVPR}, 2017.

\bibitem{zhou2016fast}
Q.-Y. Zhou, J.~Park, and V.~Koltun.
\newblock {Fast Global Registration}.
\newblock In {\em European Conference on Computer Vision}, pages 766--782.
  Springer, 2016.

\bibitem{zhou2018open3d}
Q.-Y. Zhou, J.~Park, and V.~Koltun.
\newblock {Open3D: A Modern Library for 3D Data Processing}.
\newblock {\em arXiv:1801.09847}, 2018.

\end{thebibliography}
}

\end{document}